\documentclass{article}
\usepackage{ijcai22}

\usepackage{times}
\usepackage{soul}
\usepackage{url}
\usepackage[hidelinks]{hyperref}
\usepackage[utf8]{inputenc}
\usepackage[small]{caption}
\usepackage{graphicx}
\usepackage{amsmath}
\usepackage{amsthm}
\usepackage{booktabs}
\usepackage{algorithm}
\usepackage{algorithmic}
\usepackage{bm}
\usepackage{multirow}
\usepackage{adjustbox}
\usepackage{tabularx}

\title{DictBERT: Dictionary Description Knowledge Enhanced Language Model Pre-training via Contrastive Learning}

\author{
Qianglong Chen$^{1,2}$
\and
Feng-Lin Li$^{2}$\and
Guohai Xu$^2$\and
Ming Yan$^2$\and
Ji Zhang$^2$\and
Yin Zhang$^1$\thanks{Corresponding Author: Yin Zhang.}
\affiliations
$^1$College of Computer Science and Technology, Zhejiang University, China\\
$^2$Alibaba Group, China\\
\emails
\{chenqianglong, zhangyin98\}@zju.edu.cn,
\{guohai.xgh, ym119608, zj122146\}@alibaba-inc.com,
maillifenglin@gmail.com
}

\begin{document}

\maketitle

\begin{abstract}
Although pre-trained language models (PLMs) have achieved state-of-the-art performance on various natural language processing (NLP) tasks, they are shown to be lacking in knowledge when dealing with knowledge driven tasks. Despite the many efforts made for injecting knowledge into PLMs, this problem remains open. To address the challenge, we propose \textbf{DictBERT}, a novel approach that enhances PLMs with dictionary knowledge which is easier to acquire than knowledge graph (KG). During pre-training, we present two novel pre-training tasks to inject dictionary knowledge into PLMs via contrastive learning: \textit{dictionary entry prediction} and \textit{entry description discrimination}. In fine-tuning, we use the pre-trained DictBERT as a plugin knowledge base (KB) to retrieve implicit knowledge for identified entries in an input sequence, and infuse the retrieved knowledge into the input to enhance its representation via a novel extra-hop attention mechanism. We evaluate our approach on a variety of knowledge driven and language understanding tasks, including NER, relation extraction, CommonsenseQA, OpenBookQA and GLUE. Experimental results demonstrate that our model can significantly improve typical PLMs: it gains a substantial improvement of 0.5\%, 2.9\%, 9.0\%, 7.1\% and 3.3\% on BERT-large respectively, and is also effective on RoBERTa-large.
\end{abstract}

\section{Introduction}

Pre-trained language models (PLMs) such as BERT ~\cite{devlin-etal-2019-bert}, RoBERTa~\cite{liu2019roberta} and ALBERT~\cite{lan2019albert} have been prevailing in both academic and industrial community due to their state-of-the-art performance on various natural language processing (NLP) tasks. However, as they capture only a general language representation learned from large-scale corpora, they are shown to be lacking in knowledge when dealing with knowledge driven tasks~\cite{talmor-etal-2019-commonsenseqa,mihaylov-etal-2018-suit}.
To address this challenge, many efforts, such as ERNIE-THU~\cite{zhang-etal-2019-ernie}, KEPLER~\cite{wang2021KEPLER}, KnowBERT~\cite{Peters2019KnowledgeEC}, K-Adapter~\cite{wang-etal-2021-k} and ERICA~\cite{qin-etal-2021-erica} have been made for injecting knowledge into PLMs for further improvement.

However, existing knowledge enhanced PLMs (i.e., K-PLMs) still suffer from several deficiencies. First, few methods pay attention to knowledge itself, including what type of knowledge is needed and the feasibility of acquiring such knowledge. On the one hand, some models take for granted the use of knowledge graph (KG), which is difficult to acquire in practice and shown to be less effective than dictionary knowledge
~\cite{xu-etal-2021-fusing,chen2020improving}. On the other hand, many methods use Wikipedia, which is easier to access but often noisy and of low knowledge density. Second, current K-PLMs mainly focus on one or two types of knowledge-driven tasks. Although they are shown to be useful on a few specific tasks, their language understanding ability was either not further validated on GLUE~\cite{liu2020k,wang-etal-2021-k} or not improved~\cite{zhang-etal-2019-ernie}. That is, the application scope of such K-PLMs is limited.

Inspired by the hint that dictionary knowledge can be even more effective than structured knowledge~\cite{chen2020improving}, we leverage dictionary sources as external knowledge to enhance PLMs. In our experience, this enjoys several benefits. First, it is consistent with human reading habit and cognitive process. In the process of reading, when encountering unfamiliar words, people usually consult dictionaries or encyclopedias. Second, compared with long Wikipedia texts, dictionary knowledge is more concise and of high knowledge density. Third, dictionary knowledge is much easier to access, which is of key importance for applying K-PLMs in practice. Even in the case of lacking a dictionary, it can be acquired through simply constructing a generator to summarize the description explaining a word. 

Correspondingly, we propose \textbf{DictBERT}, an effective approach that enhances PLMs with dictionary knowledge via contrastive learning. In the pre-training stage, we inject dictionary knowledge into PLMs through two novel pre-training tasks: \textit{dictionary entry prediction}, in which we use a description to predict its masked entry and learn entry representations from descriptive texts; and \textit{entry description discrimination}, where we use contrastive learning to improve the robustness of entry representations by constructing positive and negative samples with dictionary synonyms and antonyms. During fine-tuning, we first identify dictionary entries from a given input, then use DictBERT as a plugin KB to retrieve corresponding entry information. For the fusion of retrieved entry information and original input, we propose a novel extra-hop attention mechanism to enhance its representation for downstream tasks.

The main contributions of our paper are as follows:
\begin{itemize}
    \item We propose DictBERT, a novel approach enhancing PLMs with dictionary knowledge, which is able to effectively not only integrate external knowledge into but also improve the language understanding ability of PLMs.
    \item For pre-training, we present two novel pre-training tasks with contrastive learning, namely {dictionary entry prediction} and {entry description discrimination}, for injecting dictionary knowledge into PLMs. For fine-tuning, we present three knowledge infusion mechanisms to utilize the retrieved knowledge from pre-trained DictBERT for improving downstream tasks.
    \item We conducted a series of experiments on NER, relation extraction (RE), CommonsenseQA, OpenBookQA and GLUE. Experimental results show that our model can significantly improve typical PLMs (BERT-large and RoBERTa-large).
\end{itemize}

\section{Related Work}
\paragraph{Knowledge Enhanced PLMs.}
To alleviate the problem of lacking knowledge for PLMs, a popular approach is to inject factual knowledge through infusing pre-trained entity embeddings~\cite{zhang-etal-2019-ernie,Peters2019KnowledgeEC} or incorporating symbolic knowledge triples~\cite{liu2020k}. One problem of using pre-trained entity embeddings, as pointed out by CoLAKE~\cite{sun2020colake}, is the separation between entity embedding and language embedding. To tackle this problem, we use textual entry descriptions to predict their masked entries in the entry prediction pre-training task. Being different from KnowBERT~\cite{Peters2019KnowledgeEC} and KEPLER~\cite{wang2021KEPLER} that use structured KGs, we use semi-structured dictionary knowledge. Inspired by K-Adapter~\cite{wang-etal-2021-k}, we also use the PLM enhanced with dictionary knowledge as a plugin for downstream tasks.
It should be noted that Dict-BERT~\cite{yu2021dict} and our work are at the same period. There are many differences between them in pre-training and fine-tuning, and our results are superior to those of Dict-BERT. 


\paragraph{Contrastive Learning.}
The main idea of contrastive learning is to improve the robustness of representations through bringing closer positive samples and pushing away negative ones. It has been widely used to obtain better sentence representations~\cite{logeswaran2018efficient,wu2020clear,wang2021cline,qin-etal-2021-erica}. While ~\cite{logeswaran2018efficient} take a sentence B that follows A as a positive example and randomly chooses a sentence C from other documents as a negative, \cite{wang2021cline} construct positive and negative examples via replacing representative tokens in a sentence with WordNet, and~\cite{wu2020clear} present multiple sentence-level data augmentation strategies. In addition, ~\cite{qin-etal-2021-erica} leverage contrastive pre-training to improve the ability of PLMs on capturing relational facts in texts. Being differently, we use synonyms and antonyms in dictionary to construct contrastive pairs, and use contrastive pre-training to learn a better dictionary entry representation for downstream tasks.

\begin{figure*}[ht]
    \centering
    \includegraphics[width=1\linewidth]{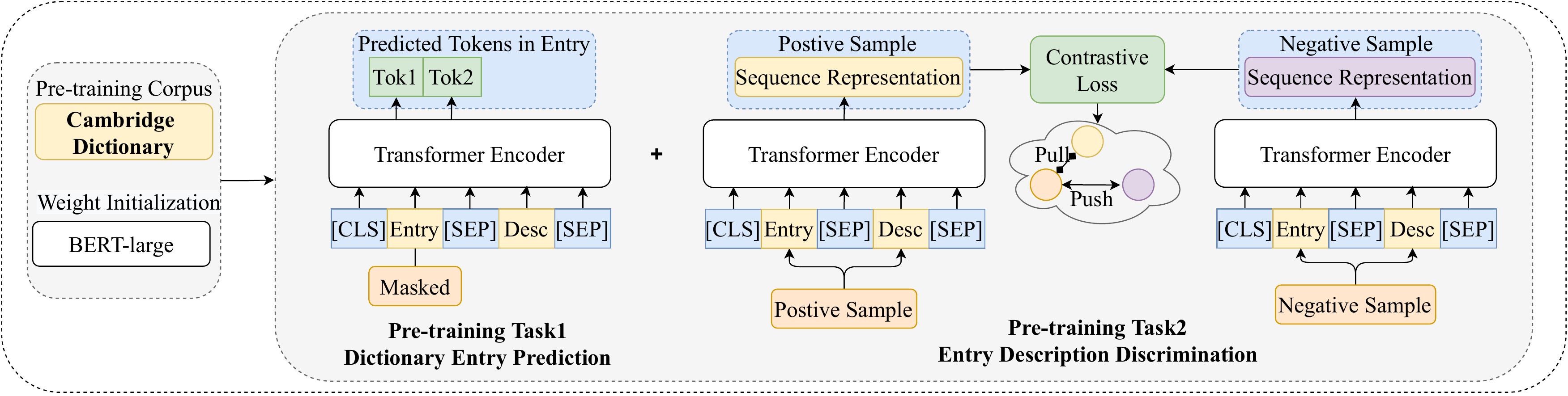}
    \caption{DictBERT pre-training. We take the Cambridge dictionary as our knowledge source. The pre-training tasks include \textbf{dictionary entry prediction} and \textbf{entry description discrimination}. In the former task, we mask only the entry tokens in the chosen input samples. In the latter task, we try to obtain better entry representations through contrastive learning.}
    \label{fig:DictBERT}
\end{figure*}

\section{The Proposed Approach}

\begin{table}[tbp]
    \centering
    \small
    \setlength{\tabcolsep}{5pt}
    \begin{tabular}{|l|l|}
    \hline
        \textbf{Entry word} & forest\\
        \hline
        \multirow{2}{*}{\textbf{Description}} & a large area of land covered with trees and plants, \\ & usually larger than a wood, or the trees and plants \\ & themselves\\
        \hline
        \textbf{Synonyms} & jungle, woodland\\
        \hline
        \textbf{Antonyms} & desert, wasteland\\
        \hline
    \end{tabular}
    \caption{A sample of dictionary entries.}
    \label{tab:dict definition}
\end{table}

\subsection{Dictionary Description Knowledge}
A dictionary is a resource that lists the words of a language, clarifies their meanings through explanatory descriptions, and often specifies their pronunciation, origin, usage, synonyms and antonyms, etc. Table~\ref{tab:dict definition} shows an example about the entry word ``forest''. 
In this paper, we use four kinds of information for pre-training: each entry, its description(s), synonym(s) and antonym(s). We leverage dictionary entry words and their meanings (i.e., explanatory descriptions) for knowledge injection pre-training. Also, in order to improve the robustness of entry representation, we use the synonyms and antonyms of an entry word for contrastive learning.

\subsection{Pre-training DictBERT}
\label{sec:preDictBERT}
As shown in Figure~\ref{fig:DictBERT}, we use two novel pre-training tasks: (1) dictionary entry prediction and (2) entry description discrimination, to capture the different aspects of dictionary knowledge through further training a PLM.
\paragraph{Dictionary Entry Prediction.}
For entry word prediction, we follow the design of masked language modeling (MLM) in BERT~\cite{devlin-etal-2019-bert}, but impose constraints on the tokens to be masked. Originally, given an input sequence, the MLM task randomly masks a certain percentage of the input tokens with a special [MASK] symbol, and then tries to recover them. Inspired by~\cite{tsukagoshi-etal-2021-defsent}, to effectively learn entry representations, we take as input the concatenation of each entry word {$e=\{t_1,...,t_i,...,t_m\}$} and its description $desc=\{w_1,w_2,...,w_n\}$ in a dictionary $D$, perform masking only on the tokens of entry $e$ in a chosen input sample $s=\{[CLS]e[SEP]desc[SEP]\}$, and at last predict the masked entry tokens based on the corresponding description $desc$. Note that if an entry $e$ consists of multiple tokens, all of the component tokens will be masked. In the case of polysemy, where an entry $e$ has multiple meanings (i.e., descriptions), we construct an input sample for each meaning in a similar way.
We formulate the entry token prediction as:
\begin{equation}
    P(t_1,...,t_i,...,t_m|s\backslash \{t_1,...,t_i,...,t_m\})
\end{equation}
where the $t_i$ is the $i$-th token of $e$, and 
$s\backslash \{t_1,...,t_i,...,t_m\}$
denotes the sample $s$ with entry tokens $t_{i...m}$ being masked.
We initialize our model with the pre-trained checkpoint of BERT-large and keep MLM as one of our objectives, which uses the cross-entropy loss as loss function $L_{dep}$.

\paragraph{Entry Description Discrimination.}
To better capture the semantics of dictionary entries, we introduce entry description discrimination, which tries to improve the robustness of entry representations through contrastive learning. Specifically, we construct positive (resp. negative) samples as follows: given an entry word $e$ and its description $desc$, we obtain its synonyms $D_s=\{e_{syn}\}$ (resp. antonyms $D_a=\{e_{ant}\}$) from the dictionary source, and treat the concatenation of each $e_{syn}$ (resp. $e_{ant}$) and its description $desc_{syn}$ (resp. $desc_{ant}$) as a positive (resp. negative) sample. Take the entry ``forest'' in Table~\ref{tab:dict definition} for example, ``woodland'' and ``desert'' are one of its synonyms and antonyms, respectively. The corresponding positive and negative samples are shown in Table~\ref{tab:cl_example}. In our experiments, we use the same number of  (e.g., 5) positive and negative samples.
Note that we currently only utilize the antonyms of an entry word to construct strict negative samples, and will explore the construction of negative samples through random selection in the future.

\begin{figure*}[!ht]
    \centering
    \includegraphics[width=0.96\linewidth]{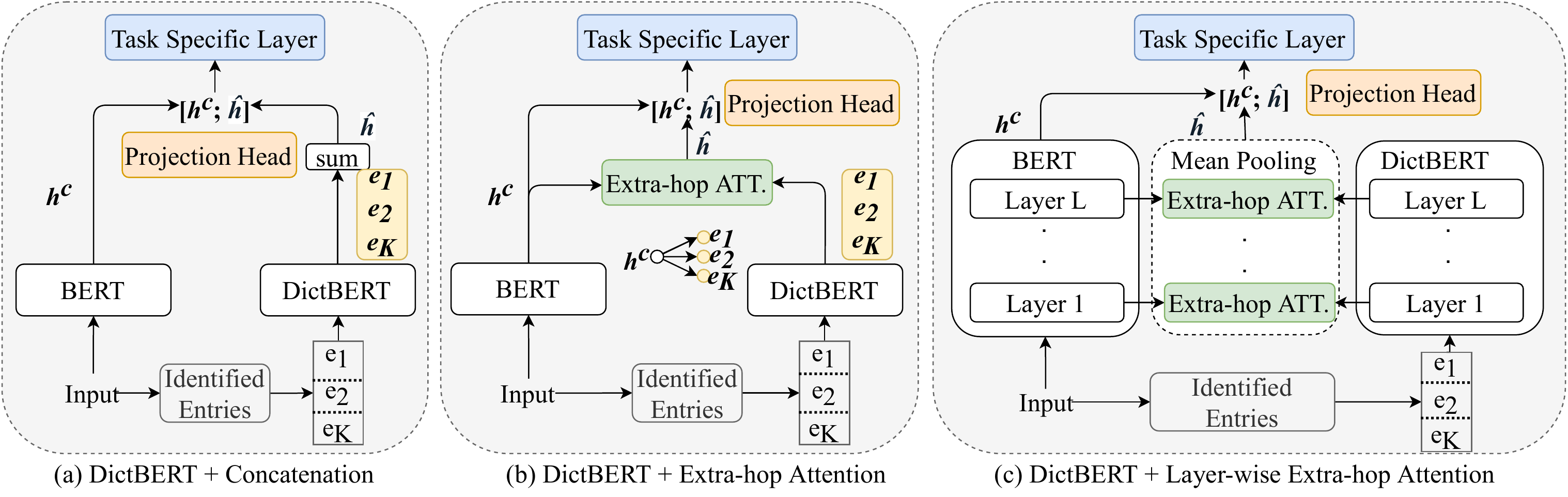}
    \caption{Downstream task fine-tuning with DictBERT, where we present three different knowledge infusion mechanisms.}
    \label{fig:Fine-tuning}
\end{figure*}

\begin{table}[tbp]
    \centering
    \small
    \setlength{\tabcolsep}{1pt}
    \begin{tabular}{|l|l|}
    \hline
    \textbf{Positive} & [CLS] woodland [SEP] Land covered with wood or trees \\& [SEP]\\
    \hline
    \textbf{Negative} & [CLS] desert [SEP] arid land with little or no vegetation \\& [SEP] \\
    \hline
    \end{tabular}
    \caption{The positive sample ``woodland'' and negative example ``desert'' of the entry ``forest''.}
    \label{tab:cl_example}
\end{table}

We use $h_{ori}$, $h_{syn}$, $h_{ant}$ to indicate the representations of the original, the positive, and the negative input sample. To bring closer $h_{ori}$ and $h_{syn}$, and push away $h_{ori}$ and $h_{ant}$, we develop a contrastive objective, where ($e_{ori}$, $e_{syn}$) is considered a positive pair and ($e_{ori}$, $e_{ant}$) is considered negative. We use $h^c$, which denotes the hidden state of the special symbol [CLS], to indicate the representation of an input sample. We define a contrastive objective $L_{edd}$: 
\begin{equation}
h^c_{ori} = Encoder(e_{ori},desc_{ori})
\end{equation}
\begin{equation}
h^c_{syn} = Encoder(e_{syn},desc_{syn})
\end{equation}
\begin{equation}
h^c_{ant} = Encoder(e_{ant},desc_{ant})
\end{equation}
\begin{equation}
    f(h^{c}_{i},h^{c}_{j} ) = exp(h^{c}_{i} h^c_{j})
\end{equation}
\begin{equation}
    L_{edd} = -\sum_{e \in D} log\frac{f(h^c_{ori},h^c_{syn})}{f(h^c_{ori},h^c_{syn}) + f(h^c_{ori},h^c_{ant})}
\end{equation}
where the $f(\bm{x},\bm{y})$ denotes the exponentiation of the dot product between hidden states $\bm{x}$ and $\bm{y}$. 

We sum the dictionary entry prediction task loss and the entry description discrimination task loss, and finally obtain the overall loss function $L$:
\begin{equation}
    L = \lambda_1 L_{dep} + \lambda_2 L_{edd}
\end{equation}
where $L_{dep}$ and $L_{edd}$ denote the loss functions of the two tasks. In our experiments, we set $\lambda_1=0.4$ and $\lambda_2=0.6$.

\subsection{Fine-tuning with DictBERT}
Inspired by ~\cite{petroni-etal-2019-language}
, we use DictBERT as a plugin with a backbone PLM during fine-tuning (i.e., it is frozen). In this way, we can enjoy the flexibility of training different DictBERTs for different dictionaries and avoid the catastrophic forgetting problem of continuous training. Specifically, we first identify dictionary entries from a given input, 
then use DictBERT as a KB to retrieve corresponding entry information (i.e., entry embeddings), and finally inject the retrieved entry information into the original input to get an enhanced representation for downstream tasks. In the case an input consists of more than one sequence (e.g., NLI), we process each input sequence individually and then feed them into the downstream task specific layer for subsequent processing. To better leverage the retrieved implicit knowledge on downstream tasks, we introduce three different kinds of knowledge infusion mechanisms (See Figure~\ref{fig:Fine-tuning}): (1) pooled output concatenation, (2) extra-hop attention and (3) layer-wise extra-hop attention. 

\paragraph{Pooled Output Concatenation.} 
As shown in Figure~\ref{fig:Fine-tuning} (a), we directly concatenate the pooled output of the backbone BERT (i.e., $h^c$) and the sum of entry embeddings retrieved from DictBERT (i.e., $\hat{h}$). Then, we feed the concatenation (i.e., [$h^c$; $\hat{h}$]) into a task specific layer for downstream tasks.

\paragraph{Extra-hop Attention.}
The simplest way to incorporate identified entries into original text is to sum up their embeddings and concatenate the summation with the text representation. However, this method can not tell which entry is more important, and which sense is more suitable in the case of polyseme entries. Therefore, we further propose an extra-hop attention mechanism to address this deficiency. 
As shown in Figure~\ref{fig:Fine-tuning} (b), we follow Transformer-XH ~\cite{Zhao2020Transformer-XH:} to use $h^c$, the hidden state of the [CLS] token in an input query as the ``attention hub'', which attends to each entry word identified in the same input. With the attentive weights, our method focuses on more important entries or meanings when integrating them as external knowledge into the original input query. The extra hop attention mechanism is formulated as:
\begin{equation}
    \hat{h} =  \sum_{i=1}^{K} ATT(h^c,e_i)
\end{equation}
where $e_i$ denotes the DictBERT output of $i_{th}$ identified entry, $K$ is the number of identified entries in the input query, and $\hat{h}$ denotes the weighted sum of retrieved entry embeddings. After we obtain the $\hat{h}$, we use [$h^c$; $\hat{h}$] for the final inference.

\paragraph{Layer-wise Extra-hop Attention.}
To further improve performance, we extend the extra-hop attention at the last layer to each inner layer, making it become layer-wise. As shown in Figure~\ref{fig:Fine-tuning} (c), we compute the attention score at each layer, and finally use their mean for implicit entry knowledge injection. Specifically, the layer-wise extra-hop attention can be formulated as:
\begin{equation}
    \hat{h}_l = \sum_{i=1}^{K} ATT(h_l,e^l_i)
\end{equation}
\begin{equation}
    \hat{h} = \frac{1}{L}\sum_{l=1}^{L} \hat{h}_l
    \label{mean}
\end{equation}
where $\hat{h}_l$ denotes the weighted sum of $l$-th layer outputs of DictBERT. With the final implicit $\hat{h}$ obtained via Equation~\ref{mean}, we use [$h^c$; $\hat{h}$] in a similar way for downstream tasks.

\section{Experiments}
\subsection{Datasets and Tasks}
\paragraph{Pre-training Dictionary Source.}
To pre-train DictBERT, we use the Cambridge Dictionary\footnote{https://dictionary.cambridge.org}, which includes 315K entry words, as our pre-training corpus. We construct input samples for the two pre-training tasks, namely dictionary entry prediction and entry description discrimination, as introduced in the section of the proposed approach.
\paragraph{CoNLL2003 \& TACRED.}
We use these two traditional knowledge-driven tasks, CoNLL2003 ~\cite{tjong-kim-sang-de-meulder-2003-introduction} and TACRED ~\cite{zhang-etal-2017-position}, to have a quick check on the effectiveness of our approach.
\paragraph{CommonsenseQA \& OpenBookQA.}
We use CommonsenseQA~\cite{talmor-etal-2019-commonsenseqa} and OpenBookQA~\cite{mihaylov-etal-2018-suit} to evaluate the ability of DictBERT acting as KBs and providing implicit knowledge to downstream tasks.
\paragraph{GLUE.}
We follow existing knowledge enhanced PLMs such as KEPLER and KnowBERT to use GLUE~\cite{wang-etal-2018-glue} to evaluate the general natural language understanding capability of our approach. 

\subsection{Experimental Settings}
For pre-training, we use the BERT-large-uncased and RoBERTa-large model as backbone and set the learning rate to $1e^{-5}$, dropout rate to $0.1$, max-length of tokens to $128$, batch size to $32$, and number of epochs to $10$. We use AdamW as the optimizer. For fine-tuning, we adopt cross-entropy loss as the loss function, set batch size to $32$ and number of epochs to $30$. We run 5 times for each task and report their average. 

\subsection{Baselines}
\paragraph{BERT \& RoBERTa.}
We adopt BERT-large~\cite{devlin-etal-2019-bert} instead of BERT-base as baseline because the former is more difficult to improve. To be more convincing, we also use the more adequately trained RoBERTa-large~\cite{liu2019roberta} for comparison in our experiments.
\paragraph{Enhanced BERT \& RoBERTa.}
For CommonsenseQA, we use BERT+AMS~\cite{ye2019align}, BERT+OMCS, RoBERTa+CSPT, RoBERTa+KE, G-DAUG~\cite{yang-etal-2020-generative} as baselines for comparison.  For OpenbookQA, we use AristoBERTv7, AristoRoBERTav7 and BERT Multi-Task as baselines for comparison.
\paragraph{KnowBERT \& KEPLER.}
For GLUE, we use KnowBERT and KEPLER as baselines for comparison.
KnowBERT~\cite{Peters2019KnowledgeEC} enhances contextual word representations through embedding structured, human-curated knowledge 
into BERT-base through entity linking and word-to-entity attention.
KEPLER~\cite{wang2021KEPLER} encodes textual entity descriptions with RoBERTa-base as their embeddings, and then jointly optimizes the knowledge embedding and language modeling objectives. 

\subsection{DictBERT Variants}
We evaluate different variants of DictBERT in our experiments.
\textbf{DictBERT+Concat(K)} uses the concatenation mechanism, \textbf{DictBERT+EHA(K)} and \textbf{DictBERT+EHA (K+V)} adopt the extra-hop attention mechanism, and \textbf{DictBERT+LWA(K+V)} uses layer-wise attention.
The symbol \textbf{K} indicates the use of entry word to retrieve entry embeddings from DictBERT, \textbf{K+V} denotes that we use both entry word and its corresponding description for knowledge retrieval. 

\begin{table}[tbp]
    \centering
    \begin{tabular}{|l|c|c|}
    \hline
        \textbf{Model} & \textbf{CoNLL2003} & \textbf{TACRED} \\
        \hline
        BERT-large & 92.8 & 70.1\\
        \hline
        DictBERT + Concat(K) & 93.1 & 72.3 \\
        DictBERT + EHA(K) & 93.2 & 72.7\\
        DictBERT + EHA(K+V) & 93.3 & 72.8\\
        DictBERT + LWA(K+V) & \textbf{93.3} & \textbf{73.0}\\
        \hline
    \end{tabular}
    \caption{Experimental results on CoNLL2003 (NER) and TACRED (relation extraction).}
    \label{tab:NER-RE}
\end{table}

\begin{table}[tbp]
    \centering
    \begin{tabular}{|l|c|c|}
    \hline
        \textbf{Model}   & \textbf{CSQA} & \textbf{OBQA}\\
        \hline
        BERT-large  & 56.7 & 60.4\\
        BERT-large + AMS  & 62.2 & -\\
        BERT-large + OMCS & 62.5 & -  \\
        BERT-large Multi-Task  & - & 63.8 \\
        AristoBERTv7-large  & 64.6 & {72.0}\\
        \hline
        RoBERTa-large  & 72.1 & 71.8\\
        RoBERTa-large + CSPT & 69.6  & -  \\
        RoBERTa-large + G-DAUG-Combo & 72.6 & -  \\
        RoBERTa-large + KE & 73.3  & - \\
        AristoRoBERTav7-large  & - & 77.8\\
        \hline
        DictBERT + Concat(K)  & 62.7 & 64.4\\
        DictBERT + EHA(K)  & 65.1 & 66.3\\
        DictBERT + EHA(K+V)  & 65.4 & 66.7\\
        DictBERT + LWA(K+V)  & {65.7}  & 67.5 \\
        \hline
        
        DictRoBERTa + Concat(K)  & 75.7 & 75.2\\
        DictRoBERTa + EHA(K) & 77.5  & 77.6\\
        DictRoBERTa + EHA(K+V)  & 77.8 & 78.1\\
        DictRoBERTa + LWA(K+V)  & \textbf{78.5} & \textbf{78.3}\\
        \hline
    \end{tabular}
    \caption{Experimental results on CommonsenseQA (CSQA) and OpenBookQA (OBQA). 
    }
    \label{tab:OpenBookQA}
\end{table}

\begin{table*}[tbp]
    \centering
    \begin{tabular}{|l|c|c|c|c|c|c|c|c|c|c|}
        \hline
         \multirow{2}{*}{\textbf{Model}}& \multicolumn{2}{c|}{\textbf{Single-sentence}} & \multicolumn{3}{c|}{\textbf{Similarity Paraphrase}} & \multicolumn{3}{c|}{\textbf{Inference}} & \multirow{2}{*}{\textbf{Avg}}\\
        \cline{2-9}
          & \textbf{CoLA} & \textbf{SST-2} & \textbf{QQP} & \textbf{STS-B}  & \textbf{MRPC} & \textbf{QNLI} & \textbf{MNLI} & \textbf{RTE} & \\
        
        \hline
        BERT-base & 52.1 & 93.5 & 88.9 & 85.8 & 88.9 & 90.5 & 84.6 & 66.4  & 81.3 \\
        BERT-large & 60.5 & 94.9 & 89.3 & 87.6  & 89.3 & 92.7 & 85.9 & 70.1 &  83.8 \\
        RoBERTa-base & 63.6 & 94.8 & 91.9 & 91.2  & 90.2 & 92.7 & 87.5 & 80.9 & 86.6 \\ 
        RoBERTa-large & 67.8 & 96.7& 90.2 & 92.0  & 93.0 & 95.4 & 90.2 & 87.2 &  89.0 \\
        \hline
        KnowBERT-WordNet+Wiki & 54.6 & 93.6&90.3 & 89.1  & 88.2 & 91.5 & 85.7 & 73.8 &  83.3\\
        KEPLER-only Desc & 55.8 & 94.4 &90.8 & 90.2  & 88.5 & 92.4 & 	85.9 & 78.3 &  84.5 \\
        KEPLER-wiki  &  63.6 & 94.5 &\textbf{91.7} & 91.2  & 89.3 & 92.4 & 87.2 & 85.2 & 86.8\\
        
        \hline

        DictBERT + Concat(K) & 64.1 & 95.4& 90.3 & 90.5  & 91.8 & 95.1 &	87.1 & 75.2	& 86.2 \\
        DictBERT + EHA(K) & 64.2  & 95.6&90.3 & 91.5 & 92.0 &	95.7  & 88.4 & 76.6 &  86.8\\
        DictBERT + EHA(K+V) & 64.5 & 95.6 &90.4 & 91.6  & 92.1 & 95.8 &88.5 & 76.8 &  86.9 \\
        DictBERT + LWA(K+V) &  64.7 &95.7 &90.4 & 91.7  & 92.3 & 96.0 & 88.7& 77.1 & 87.1 \\
        \hline
        DictRoBERTa + Concat(K) & 68.1 & 97.1 &90.3 & 91.2  & 92.5 & 96.2 & 90.5 & 88.1 &  89.2 \\
        DictRoBERTa + EHA(K) & 68.3 & 97.3 &90.4 & 91.7  & 92.7 & 96.4 & 90.8 & 88.6 &  89.5 \\
        DictRoBERTa + EHA(K+V)	& 68.5 & 97.5 &90.5 & 91.8  & 92.8 & 96.5 & 90.9 & 89.1 &  89.7 \\
        DictRoBERTa + LWA(K+V) & \textbf{68.6} & \textbf{97.8} & 90.8 & \textbf{92.1}  & \textbf{93.2} & \textbf{96.8} & \textbf{91.1} & \textbf{89.4} & \textbf{89.9} \\
        \hline

    \end{tabular}
    \caption{Experimental results on the GLUE development set. The parameter of DictBERT is based on BERT-large. For parameter initialization, KnowBERT uses the BERT-base, while KEPLER uses RoBERTa-base.}
    \label{tab:GLUE}
\end{table*}

\subsection{Experimental Results and Analysis}
\paragraph{Traditional Knowledge Driven Task Results.} 

Firstly, we evaluate DictBERT on NER and relation extraction, the most commonly used knowledge driven tasks. As shown in Table~\ref{tab:NER-RE}, our approach is finally able to improve the performance on CoNLL2003 and TACRED by 0.5\% and 2.9\% compared with the strong baseline BERT-large. Further, we can observe that all the three knowledge infusion mechanisms are helpful, and the layer-wise attention achieve the best results. This indicates the identification and explanation of important entities in an input sample are of key importance. Meanwhile, we found that the use of additional entry description (i.e., the K+V setting) can help retrieve better entry embeddings.

\paragraph{Knowledge Driven QA Task Results.} 
We further assess DictBERT on knowledge driven QA tasks, namely CommonsenseQA and OpenBookQA, and report the results in Table~\ref{tab:OpenBookQA}. Compared with BERT-large, our basic setting DictBERT+Concat gains a significant improvement of 6.0\% and 4.0\% on the two tasks, respectively. Further, we observe that the extra-hop attention brings an evident increase (2.4\% and 1.9\%), verifying again the importance of identifying attentive weights of entries in an input sample. Lastly, DictBERT+LWA(K+V) achieves the best result on both tasks, bringing a final gain of 9.0\% and 7.1\% compared to the BERT-large baseline. To be more convincing, we also compare DictRoBERTa with the original RoBERTa-large on CommonsenseQA and OpenBookQA. As shown in Table~\ref{tab:OpenBookQA}, the conclusion also holds for RoBERTa. Similarly, DictRoBERTa+LWA(K+V) achieves the best results, which can ultimately improve over 6.4\% and 6.5\%, respectively.


\paragraph{GLUE Results.}
We also evaluate DictBERT on GLUE to examine whether it can improve the general natural language understanding ability of PLMs. Table~\ref{tab:GLUE} shows that compared with BERT-large our basic setting DictBERT+Concat achieves an average improvement of 2.4\%, indicating the effectiveness of injecting dictionary knowledge for language understanding. Similarly, the extra-hop attention and the use of additional entry description (i.e., the K+V setting) contribute to further improvement, and DictBERT+LWA(K+V) achieves the best results, bringing a final increase of 3.3\% on average. 
With the baseline being RoBERTa-large, our best model can achieve an average increase of 0.9\% on GLUE, validating the effectiveness and broad applicability of our approach.
As for other K-PLMs, KnowBERT enhanced with WordNet and 470K Wikipedia entities can improve BERT-base by 2.0\%, which is smaller than the performance gain (3.3\%) brought by our method on BERT-large. KEPLER-wiki can only improve RoBERTa-base by 0.2\% when using 5M Wikidata entities for knowledge (entity) embedding and extra 13GB text data for MLM. With the only 5M entity descriptions for MLM, there is an obvious performance drop for KEPLER-OnlyDesc.
Therefore, our approach is more effective in improving language understanding ability with external knowledge (we use only 315K dictionary entries).

\begin{table}[htb]
    \centering
    \setlength{\tabcolsep}{1pt}
    \begin{tabular}{|l|c|c|c|}
    \hline
        \textbf{Model}  & \textbf{CSQA} & \textbf{OBQA} & \textbf{GLUE} \\
        \hline
        BERT-large   & 56.7 & 60.4 & 83.8\\
        BERT-large + Concat(K)   & 57.1 & 60.6 & 83.9\\
        BERT-large + LWA(K+V)    & 57.3 & 60.7 & 84.1\\
        \hline
        \multicolumn{4}{|c|}{\textbf{Pre-training}}\\
        \hline
        BERT-large (dict corpus)+Concat(K) & {61.9} & {63.8}  & {85.5} \\
        DictBERT(DEP)+Concat(K)  & {62.2} & {64.1} & {85.8}\\
        DictBERT(DEP+EDD)+Concat(K) & 62.7 & 64.4 & 86.2\\
        \hline
        \multicolumn{4}{|c|}{\textbf{Fine-tuning}}\\
        \hline
        DictBERT-only  & 62.6 & 64.1 & 85.7 \\
        DictBERT + Concat(K)   & 62.7 & 64.4 & 86.2\\
        DictBERT + EHA(K)  & 65.1 & 66.3 & 86.8 \\
        DictBERT + EHA(K+V)   & 65.4 & 66.7 & 86.9\\
        DictBERT + LWA(K+V)  & 65.7 & 67.5 & 87.1 \\
        DictBERT plus  + LWA(K+V)  & {\textbf{66.1}}& {\textbf{67.7}} & {\textbf{87.2}}\\
        \hline
        
        
    \end{tabular}
    \caption{Ablation study results on CSQA, OBQA and GLUE.}
    \label{tab:Ablation_Study}
\end{table}

\paragraph{Ablation Study.}
We perform ablation studies on the different components of DictBERT. 
Firstly, we evaluate {BERT-large+Concat(K)} and {BERT-large+LWA(K+V)}, which directly use BERT-large, instead of our pre-trained DictBERT, as the plugin. As we can see, the improvement is rather marginal, confirming the necessity of injecting external knowledge. 
Secondly, we assess the effectiveness of the two pre-training tasks: DictBERT(DEP)+Concat and DictBERT(DEP+EDD)+Concat. 
As shown in Table~\ref{tab:Ablation_Study}, contrastive learning is helpful to some degree (0.4\% on average), and masking only entry tokens is better than masking tokens of both entries and descriptions (+0.3\% for all the three).
Finally, we examine the necessity of using DictBERT as a plugin KB instead of directly using it for downstream task fine-tuning (DictBERT-only), and whether the dictionary size matters (DictBERT plus). 
As shown in Table~\ref{tab:Ablation_Study}, all of our three knowledge infusion mechanisms can further improve the performance of DictBERT-only, indicating the use of DictBERT as a plugin is rewarding. 
To assess the effect of dictionary size, we use the union of the Cambridge Dictionary, the Oxford Dictionary and the Wiktionary, which totals more than 1M unique entry words. The results show that DictBERT plus+LWA(K+V) can further improve the performance of the three task sets (+0.23\% on average).

\subsection{Discussions}
Experimental results show that DictBERT can not only integrate external knowledge into but also improve the language understanding ability of PLMs. It is worth mentioning that DictBERT is assessed on very strong baselines (BERT-large and RoBERTa-large, rather than the base counterparts adopted by many other K-PLMs), which indicates the effectiveness of our method from another side. Last but not least, our approach can be easily applied in practice: dictionary source is relatively easy to acquire, through either crawling or simple generative models. As for computation cost, DictBERT as a KB plugin can be further simplified to be DictBERT as a lookup table, with each entry in a dictionary being mapped to an embedding in advance, largely accelerating the inference speed. Through generating dictionary entry embeddings in advance by using the plugin, the complexity of our approach is similar to that of the backbone PLM.

\section{Conclusion}
In this paper, we propose DictBERT, an effective approach that enhances PLMs with dictionary knowledge through two novel pre-training tasks and an attention-based knowledge infusion mechanism during downstream task fine-tuning. We also demonstrate its effectiveness through an adequate set of experiments. Importantly, our approach can be easily applied in practice. In the future, we are going to further explore more effective pre-training tasks and knowledge infusion mechanisms for injecting knowledge into multilingual pre-trained language models.

\section*{Acknowledgments}
We thank the anonymous reviewers for their helpful comments on this paper. This work was supported by National Key R\&D Program of China (No.~2018AAA0101900), the NSFC projects (No.~62072399, No.~U19B2042, No.~61402403), Chinese Knowledge Center for Engineering Sciences and Technology, MoE Engineering Research Center of Digital Library, Alibaba Group, Alibaba-Zhejiang University Joint Research Institute of Frontier Technologies, and the Fundamental Research Funds for the Central Universities (No.~226-2022-00070).

\bibliographystyle{named}
\bibliography{ijcai22}

\begin{thebibliography}{}

\bibitem[\protect\citeauthoryear{Chen \bgroup \em et al.\egroup
  }{2020}]{chen2020improving}
Qianglong Chen, Feng Ji, Haiqing Chen, and Yin Zhang.
\newblock Improving commonsense question answering by graph-based iterative
  retrieval over multiple knowledge sources.
\newblock In {\em COLING}, pages 2583--2594, 2020.

\bibitem[\protect\citeauthoryear{Devlin \bgroup \em et al.\egroup
  }{2019}]{devlin-etal-2019-bert}
Jacob Devlin, Ming-Wei Chang, Kenton Lee, and Kristina Toutanova.
\newblock {BERT}: Pre-training of deep bidirectional transformers for language
  understanding.
\newblock In {\em NAACL-HLT}, pages 4171--4186, 2019.

\bibitem[\protect\citeauthoryear{Lan \bgroup \em et al.\egroup
  }{2019}]{lan2019albert}
Zhenzhong Lan, Mingda Chen, Sebastian Goodman, Kevin Gimpel, Piyush Sharma, and
  Radu Soricut.
\newblock Albert: A lite bert for self-supervised learning of language
  representations.
\newblock {\em arXiv:1909.11942}, 2019.

\bibitem[\protect\citeauthoryear{Liu \bgroup \em et al.\egroup
  }{2019}]{liu2019roberta}
Yinhan Liu, Myle Ott, Naman Goyal, Jingfei Du, Mandar Joshi, Danqi Chen, Omer
  Levy, Mike Lewis, Luke Zettlemoyer, and Veselin Stoyanov.
\newblock Roberta: A robustly optimized bert pretraining approach.
\newblock {\em arXiv:1907.11692}, 2019.

\bibitem[\protect\citeauthoryear{Liu \bgroup \em et al.\egroup
  }{2020}]{liu2020k}
Weijie Liu, Peng Zhou, Zhe Zhao, et~al.
\newblock K-bert: Enabling language representation with knowledge graph.
\newblock In {\em AAAI}, pages 2901--2908, 2020.

\bibitem[\protect\citeauthoryear{Logeswaran and
  Lee}{2018}]{logeswaran2018efficient}
Lajanugen Logeswaran and Honglak Lee.
\newblock An efficient framework for learning sentence representations.
\newblock In {\em ICLR}, 2018.

\bibitem[\protect\citeauthoryear{Mihaylov \bgroup \em et al.\egroup
  }{2018}]{mihaylov-etal-2018-suit}
Todor Mihaylov, Peter Clark, Tushar Khot, and Ashish Sabharwal.
\newblock Can a suit of armor conduct electricity? a new dataset for open book
  question answering.
\newblock In {\em EMNLP}, pages 2381--2391, 2018.

\bibitem[\protect\citeauthoryear{Peters \bgroup \em et al.\egroup
  }{2019}]{Peters2019KnowledgeEC}
Matthew~E. Peters, Mark Neumann, Robert~L Logan, Roy Schwartz, Vidur Joshi,
  Sameer Singh, and Noah~A. Smith.
\newblock Knowledge enhanced contextual word representations.
\newblock In {\em EMNLP}, pages 43--54, 2019.

\bibitem[\protect\citeauthoryear{Petroni \bgroup \em et al.\egroup
  }{2019}]{petroni-etal-2019-language}
Fabio Petroni, Tim Rockt{\"a}schel, Sebastian Riedel, Patrick Lewis, Anton
  Bakhtin, Yuxiang Wu, and Alexander Miller.
\newblock Language models as knowledge bases?
\newblock In {\em EMNLP}, pages 2463--2473, 2019.

\bibitem[\protect\citeauthoryear{Qin \bgroup \em et al.\egroup
  }{2021}]{qin-etal-2021-erica}
Yujia Qin, Yankai Lin, Ryuichi Takanobu, Zhiyuan Liu, Peng Li, Heng Ji, Minlie
  Huang, Maosong Sun, and Jie Zhou.
\newblock {ERICA}: Improving entity and relation understanding for pre-trained
  language models via contrastive learning.
\newblock In {\em ACL}, pages 3350--3363, 2021.

\bibitem[\protect\citeauthoryear{Sun \bgroup \em et al.\egroup
  }{2020}]{sun2020colake}
Tianxiang Sun, Yunfan Shao, Xipeng Qiu, Qipeng Guo, Yaru Hu, Xuanjing Huang,
  and Zheng Zhang.
\newblock Colake: Contextualized language and knowledge embedding.
\newblock In {\em COLING}, pages 3660--3670, 2020.

\bibitem[\protect\citeauthoryear{Talmor \bgroup \em et al.\egroup
  }{2019}]{talmor-etal-2019-commonsenseqa}
Alon Talmor, Jonathan Herzig, Nicholas Lourie, and Jonathan Berant.
\newblock {C}ommonsense{QA}: A question answering challenge targeting
  commonsense knowledge.
\newblock In {\em NAACL-HLT}, pages 4149--4158, 2019.

\bibitem[\protect\citeauthoryear{Tjong Kim~Sang and
  De~Meulder}{2003}]{tjong-kim-sang-de-meulder-2003-introduction}
Erik~F. Tjong Kim~Sang and Fien De~Meulder.
\newblock Introduction to the {C}o{NLL}-2003 shared task: Language-independent
  named entity recognition.
\newblock In {\em NAACL-HLT}, pages 142--147, 2003.

\bibitem[\protect\citeauthoryear{Tsukagoshi \bgroup \em et al.\egroup
  }{2021}]{tsukagoshi-etal-2021-defsent}
Hayato Tsukagoshi, Ryohei Sasano, and Koichi Takeda.
\newblock {D}ef{S}ent: Sentence embeddings using definition sentences.
\newblock In {\em ACL}, 2021.

\bibitem[\protect\citeauthoryear{Wang \bgroup \em et al.\egroup
  }{2018}]{wang-etal-2018-glue}
Alex Wang, Amanpreet Singh, Julian Michael, Felix Hill, Omer Levy, and Samuel
  Bowman.
\newblock {GLUE}: A multi-task benchmark and analysis platform for natural
  language understanding.
\newblock In {\em EMNLP}, 2018.

\bibitem[\protect\citeauthoryear{Wang \bgroup \em et al.\egroup
  }{2021a}]{wang2021cline}
Dong Wang, Ning Ding, Piji Li, and Hai-Tao Zheng.
\newblock Cline: Contrastive learning with semantic negative examples for
  natural language understanding.
\newblock In {\em ACL}, pages 2332--2342, 2021.

\bibitem[\protect\citeauthoryear{Wang \bgroup \em et al.\egroup
  }{2021b}]{wang-etal-2021-k}
Ruize Wang, Duyu Tang, Nan Duan, Zhongyu Wei, Xuanjing Huang, Jianshu Ji,
  Guihong Cao, Daxin Jiang, and Ming Zhou.
\newblock {K-Adapter}: {I}nfusing {K}nowledge into {P}re-{T}rained {M}odels
  with {A}dapters.
\newblock In {\em Findings of ACL}, pages 1405--1418, 2021.

\bibitem[\protect\citeauthoryear{Wang \bgroup \em et al.\egroup
  }{2021c}]{wang2021KEPLER}
Xiaozhi Wang, Tianyu Gao, Zhaocheng Zhu, Zhengyan Zhang, Zhiyuan Liu, Juanzi
  Li, and Jian Tang.
\newblock Kepler: A unified model for knowledge embedding and pre-trained
  language representation.
\newblock {\em TACL}, 2021.

\bibitem[\protect\citeauthoryear{Wu \bgroup \em et al.\egroup
  }{2020}]{wu2020clear}
Zhuofeng Wu, Sinong Wang, Jiatao Gu, Madian Khabsa, Fei Sun, and Hao Ma.
\newblock Clear: Contrastive learning for sentence representation.
\newblock {\em arXiv:2012.15466}, 2020.

\bibitem[\protect\citeauthoryear{Xu \bgroup \em et al.\egroup
  }{2021}]{xu-etal-2021-fusing}
Yichong Xu, Chenguang Zhu, Ruochen Xu, Yang Liu, Michael Zeng, and Xuedong
  Huang.
\newblock Fusing context into knowledge graph for commonsense question
  answering.
\newblock In {\em Findings of ACL}, pages 1201--1207, 2021.

\bibitem[\protect\citeauthoryear{Yang \bgroup \em et al.\egroup
  }{2020}]{yang-etal-2020-generative}
Yiben Yang, Chaitanya Malaviya, Jared Fernandez, Swabha Swayamdipta, Ronan
  Le~Bras, Ji-Ping Wang, Chandra Bhagavatula, Yejin Choi, and Doug Downey.
\newblock Generative data augmentation for commonsense reasoning.
\newblock In {\em Findings of EMNLP}, pages 1008--1025, 2020.

\bibitem[\protect\citeauthoryear{Ye \bgroup \em et al.\egroup
  }{2019}]{ye2019align}
Zhi-Xiu Ye, Qian Chen, Wen Wang, and Zhen-Hua Ling.
\newblock Align, mask and select: A simple method for incorporating commonsense
  knowledge into language representation models.
\newblock {\em arXiv:1908.06725}, 2019.

\bibitem[\protect\citeauthoryear{Yu \bgroup \em et al.\egroup
  }{2021}]{yu2021dict}
Wenhao Yu, Chenguang Zhu, Yuwei Fang, Donghan Yu, Shuohang Wang, Yichong Xu,
  Michael Zeng, and Meng Jiang.
\newblock Dict-bert: Enhancing language model pre-training with dictionary.
\newblock {\em arXiv:2110.06490}, 2021.

\bibitem[\protect\citeauthoryear{Zhang \bgroup \em et al.\egroup
  }{2017}]{zhang-etal-2017-position}
Yuhao Zhang, Victor Zhong, Danqi Chen, Gabor Angeli, and Christopher~D.
  Manning.
\newblock Position-aware attention and supervised data improve slot filling.
\newblock In {\em EMNLP}, pages 35--45, 2017.

\bibitem[\protect\citeauthoryear{Zhang \bgroup \em et al.\egroup
  }{2019}]{zhang-etal-2019-ernie}
Zhengyan Zhang, Xu~Han, Zhiyuan Liu, Xin Jiang, Maosong Sun, and Qun Liu.
\newblock {ERNIE}: Enhanced language representation with informative entities.
\newblock In {\em ACL}, pages 1441--1451, 2019.

\bibitem[\protect\citeauthoryear{Zhao \bgroup \em et al.\egroup
  }{2020}]{Zhao2020Transformer-XH:}
Chen Zhao, Chenyan Xiong, Corby Rosset, Xia Song, Paul Bennett, and Saurabh
  Tiwary.
\newblock Transformer-xh: Multi-evidence reasoning with extra hop attention.
\newblock In {\em ICLR}, 2020.

\end{thebibliography}

\appendix
\section{More details of DictBERT}
For the DEP (dictionary entry prediction) task, we construct training examples as follows if an entry has multiple senses: ``Entry$_{1-1}$\#Description$_1$'', ``Entry$_{1-2}$\#Description$_2$''. For the EDD (entry description discrimination) task, we constrain the number of positive and negative examples, and adopt up-sampling (resp. down-sampling) if an entry has less (resp. more) synonyms or antonyms. 

Supposing that we found 3 entries $e_1$, $e_2$ and $e_3$ in an input, and $e_3$ has 2 senses $e_{3-1}$ and $e_{3-2}$ , we will end up with K = 4 entries ($e_1$, $e_2$, $e_{3-1}$, $e_{3-2}$). We next feed the 4 entries into DictBERT to get their embeddings ($e_{3-1}$ and $e_{3-2}$ have different embeddings). Taking the text [CLS] representation of the backbone PLM as $q$, and the embeddings of entries ([CLS] hidden state of DictBERT output) as $k$ and $v$, we calculate their weighted sum, and concatenate the summation to the text representation for downstream tasks. With the attentive weights, we can focus on the more important entries or meanings when integrating them as external knowledge into the original input query.

\section{More Discussion}
\paragraph{Difference from other K-PLMs}
We performed an investigation on the knowledge sources and downstream tasks of K-PLMs, and found that the majority of K-PLMs uses Wikipedia and Wikidata (and other structured KGs) as their knowledge sources, and focuses on entity-centric tasks, including sequence labeling (e.g., NER, POS), entity typing, relation classification. Only a few approaches are evaluated on question answering (QA) tasks. That is, the scope of application of such methods can be limited in practice, where QA and language understanding tasks play a key role. Our approach is designed not only for traditional knowledge driven tasks, but also for knowledge driven QA tasks and general language understanding tasks.
\paragraph{The effectiveness of plugin knowledge injection} The essence of our DictBERT plugin is to provide knowledge embeddings for dictionary entries identified in a text sequence, and that of the extra-hop attention mechanism is to contextualise the embeddings.
In the ablation study, we have evaluated: 1) directly using another BERT-large as a plugin (row 3 of table 6); 2) training a BERT-large plugin with MLM on more dictionary related training data (row 6 of table 6). As we can see from the results, these two methods only have marginal gains, suggesting the necessity and effectiveness of our approach.
To speed up training and inference, and also improve the compatibility with other techniques, we can simplify our approach through calculating the embeddings for all the entries in a dict in advance and keep them in a lookup table as our DictBERT plugin is frozen during fine-tuning. Given a text sequence, what we need to do is to identify entry words from it, retrieve the embeddings from the lookup table, and integrate them through extra-hop attention. 

\paragraph{Reasons of using large version instead of base version}
The base versions of PLMs, especially BERT-base, are often less adequately trained, and hence are easier to improve than their large versions. In practice, we found that many techniques are useful on BERT-base, but have very marginal improvement on BERT-large. We choose to focus on the harder version of PLMs, BERT-large and RoBERTa-large, to evaluate the effectiveness of our approach.

\begin{figure}
    \centering
    \includegraphics[width=1\linewidth]{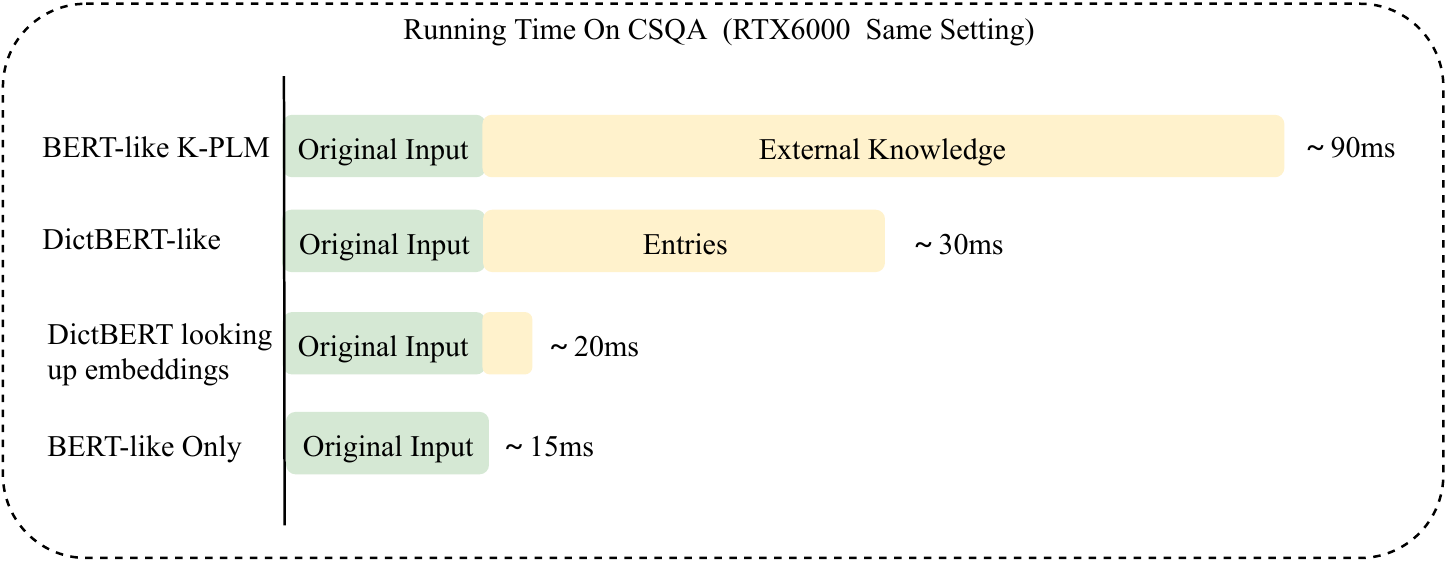}
    \caption{Running Time Comparison in Inference.}
    \label{fig:RUN_time}
\end{figure}

\paragraph{Computational costs}
The framework proposed in our paper do will increase complexity in some way, but we have accordingly proposed a simplification method.
Through generating dictionary entry embeddings in advance by using the plugin, the complexity of our approach is similar to the backbone PLM. As shown in Figure\ref{fig:RUN_time}, with the external knowledge enhanced with BERT encoder, due to the length of external knowledge, it will take 90ms in one batch. Since we only need retrieve entries from DictBERT, it can save about 60ms. Furthermore, with the simplification method, we use the lookup table for entries embedding retrieval, we can save about 10ms.

\section{Case study}
\begin{table}[htb]
    \centering
    \begin{adjustbox}{max width=0.48\textwidth}
    \begin{tabular}{|l|l|}
    \hline
        \multirow{2}{*}{Question} & What happens when someone is \\ & resting  when they are tired? \\ 
    \hline
        \multirow{2}{*}{Choices}&  A.time passes B.fall asleep C. going \\ & to sleep  D.lying down E.snore\\
    \hline
        True Label & \textbf{B. fall asleep}\\
    \hline
        BERT-large & {E.snore} \\ 
        
    \hline
        DictBERT-only & {C.going to sleep} \\
        
    \hline
        \multirow{2}{*}{DictBERT+LWA(K+V)} & \textbf{B. fall asleep} \\
        &(Entries: resting, fall asleep)\\
    \hline
    \end{tabular}
    \end{adjustbox}
    \caption{Case study of DictBERT in CommonsenseQA.}
    \label{tab:case_study}
\end{table}

\paragraph{Case Study}
In this section, we use an example in CommonsenseQA to demonstrate how our model can utilize external knowledge for question answering. As shown in Table~\ref{tab:case_study}, for the given example question ``What happens when someone is resting when they are tired?'', the baseline BERT-large predicts the distractor ``snore'', our DictBERT-only model is able to choose ``going to sleep'', which is more close. Lastly, our DictBERT+LWA(K+V) model is able to predict the correct choice ``fall asleep'' with the extra information of the entries ``resting'' and ``fall aslepp'' identified from the input\footnote{We use the concatenation of original question and each candidate as a revised input, then feed it into model for prediction.}. In general, the predictions of the two DictBERT variants are more relevant to ``sleep'', the keyword in our correct answer, indicating that injecting knowledge into PLMs is helpful.

\end{document}